%% file: main.tex
\documentclass{egpubl}
\usepackage{egsgp2023}
 
\SpecialIssuePaper         

\CGFccby

\usepackage[T1]{fontenc}
\usepackage{dfadobe}  

\biberVersion
\BibtexOrBiblatex
\usepackage[backend=biber,bibstyle=EG,citestyle=alphabetic,backref=true]{biblatex} 
\electronicVersion
\PrintedOrElectronic

\ifpdf \usepackage[pdftex]{graphicx} \pdfcompresslevel=9
\else \usepackage[dvips]{graphicx} \fi

\usepackage{egweblnk} 

\usepackage{multirow}
\usepackage{adjustbox}
\usepackage{booktabs}
\usepackage{amsmath}

\usepackage{amssymb}
\usepackage{manyfoot}
\SetFootnoteHook{\hspace*{-0.5em}}
\DeclareNewFootnote{bl}[gobble]
\usepackage{relsize}

\title[Cross-Shape Attention for Part Segmentation of 3D Point Clouds]%
      {Cross-Shape Attention for Part Segmentation of 3D Point Clouds}

\author[M.\ Loizou et al.]
{\parbox{\textwidth}{\centering Marios Loizou$^{\dagger}$$^{1}$\orcid{0000-0002-2920-0087}, Siddhant Garg$^{\dagger}$$^{2}$\orcid{0009-0004-0219-511X}, Dmitry 
Petrov\thanks{Equal contribution.}$^{2}$\orcid{0000-0003-0445-3923}, Melinos Averkiou$^{1}$\orcid{0000-0003-1814-7134} and Evangelos Kalogerakis$^{2}$\orcid{https://orcid.org/0000-0002-5867-5735}}
        \\
{\parbox{\textwidth}{\centering $^1$University of Cyprus/CYENS CoE ~~~~$^2$University of Massachusetts Amherst}
}
}

\begin{document}

\include{symbols_format}

\teaser{
\vspace*{-10mm}
    \includegraphics[width=\textwidth]{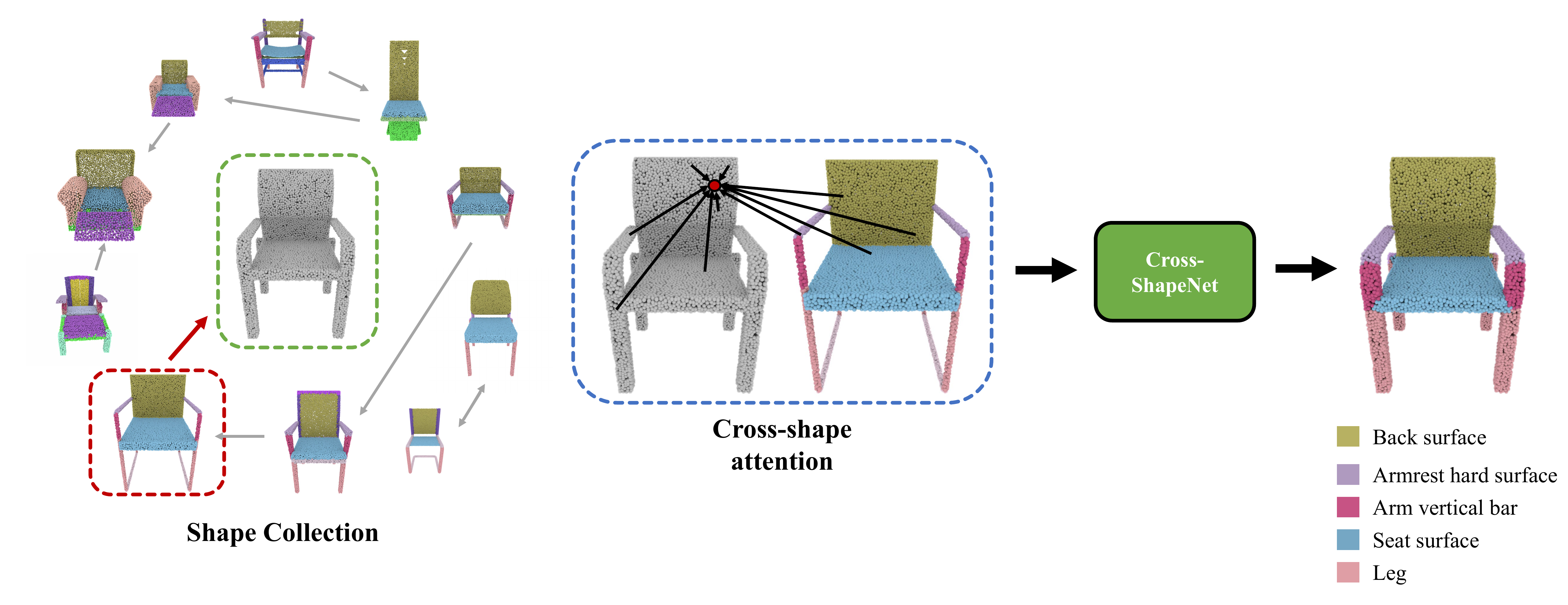}
    \centering
    \vspace{-8.5mm} 
    \caption{\emph{Left:} Given an input shape collection, our method constructs a graph where each shape is represented as a node and edges indicate shape pairs that are deemed compatible for cross-shape feature
    propagation. \emph{Middle:} 
    Our network is designed to compute point-wise feature representations for a given shape (grey shape) by enabling interactions between its own point-wise features and those of other shapes using our cross-shape attention mechanism.
    \emph{Right:}      
    As a result, the point-wise
    features of the shape become more synchronized with ones of other relevant shapes leading to more
    accurate fine-grained segmentation.    
    }
   
    \label{fig:csn_teaser}
}

\maketitle

\input{content/sections/abstract}

\footnotebl{$^{\mathlarger{\ddagger}} \ \ $This is the author’s version of the work. It is posted here for your pers\-onal use. The definitive version of the article will be published at Computer Graphics Forum, vol.42, no.5, 2023, https://doi.org/10.1111/cgf.14909.}

\input{content/sections/introduction}

\input{content/sections/related_work}

\input{content/sections/method}

\input{content/sections/results}

\input{content/sections/conclusion}

\printbibliography                

\input{supplementary.tex}

\end{document}

%% file: symbols_format.tex
\newcommand{\rev}[1]{{{#1}}} 

\newcommand{\ba}{\mathbf{a}}
\newcommand{\bb}{\mathbf{b}}
\newcommand{\bc}{\mathbf{c}}
\newcommand{\bd}{\mathbf{d}}
\newcommand{\be}{\mathbf{e}}
\newcommand{\bff}{\mathbf{f}}
\newcommand{\bg}{\mathbf{g}}
\newcommand{\bh}{\mathbf{h}}
\newcommand{\bi}{\mathbf{i}}
\newcommand{\bj}{\mathbf{j}}
\newcommand{\bk}{\mathbf{k}}
\newcommand{\bl}{\mathbf{l}}
\newcommand{\bm}{\mathbf{m}}
\newcommand{\bn}{\mathbf{n}}
\newcommand{\bo}{\mathbf{o}}
\newcommand{\bp}{\mathbf{p}}
\newcommand{\bq}{\mathbf{q}}
\newcommand{\br}{\mathbf{r}}
\newcommand{\bs}{\mathbf{s}}
\newcommand{\bt}{\mathbf{t}}
\newcommand{\bu}{\mathbf{u}}
\newcommand{\bv}{\mathbf{v}}
\newcommand{\bw}{\mathbf{w}}
\newcommand{\bx}{\mathbf{x}}
\newcommand{\by}{\mathbf{y}}
\newcommand{\bz}{\mathbf{z}}
\newcommand{\bA}{\mathbf{A}}
\newcommand{\bB}{\mathbf{B}}
\newcommand{\bC}{\mathbf{C}}
\newcommand{\bD}{\mathbf{D}}
\newcommand{\bE}{\mathbf{E}}
\newcommand{\bF}{\mathbf{F}}
\newcommand{\bG}{\mathbf{G}}
\newcommand{\bH}{\mathbf{H}}
\newcommand{\bI}{\mathbf{I}}
\newcommand{\bJ}{\mathbf{J}}
\newcommand{\bK}{\mathbf{K}}
\newcommand{\bL}{\mathbf{L}}
\newcommand{\bM}{\mathbf{M}}
\newcommand{\bN}{\mathbf{N}}
\newcommand{\bO}{\mathbf{O}}
\newcommand{\bP}{\mathbf{P}}
\newcommand{\bQ}{\mathbf{Q}}
\newcommand{\bR}{\mathbf{R}}
\newcommand{\bS}{\mathbf{S}}
\newcommand{\bT}{\mathbf{T}}
\newcommand{\bU}{\mathbf{U}}
\newcommand{\bV}{\mathbf{V}}
\newcommand{\bW}{\mathbf{W}}
\newcommand{\bX}{\mathbf{X}}
\newcommand{\bY}{\mathbf{Y}}
\newcommand{\bZ}{\mathbf{Z}}
\newcommand{\balpha}{\mbox{\boldmath$\alpha$}}
\newcommand{\bgamma}{\mbox{\boldmath$\gamma$}}
\newcommand{\bGamma}{\mbox{\boldmath$\Gamma$}}
\newcommand{\bmu}{\mbox{\boldmath$\mu$}}
\newcommand{\bphi}{\mbox{\boldmath$\phi$}}
\newcommand{\bPhi}{\mbox{\boldmath$\Phi$}}
\newcommand{\bSigma}{\mbox{\boldmath$\Sigma$}}
\newcommand{\bsigma}{\mbox{\boldmath$\sigma$}}
\newcommand{\btheta}{\mbox{\boldmath$\theta$}}

\newcommand{\mE}{\mathcal{E}}
\newcommand{\mV}{\mathcal{V}}
\newcommand{\mM}{\mathcal{M}}
\newcommand{\mH}{\mathcal{H}}
\newcommand{\mL}{\mathcal{L}}
\newcommand{\mU}{\mathcal{U}}
\newcommand{\mC}{\mathcal{C}}
\newcommand{\mS}{\mathcal{S}}
\newcommand{\mR}{\mathcal{R}}
\newcommand{\mD}{\mathcal{D}}
\newcommand{\mO}{\mathcal{O}}
\newcommand{\mP}{\mathcal{P}}
\newcommand{\mT}{\mathcal{T}}
\newcommand{\mSl}{\mathcal{S}_l}
\newcommand{\mN}{\mathcal{N}}
\newcommand{\mDll}{\mathcal{D}_{l,l'}}

\newcommand{\ra}{\rightarrow}
\newcommand{\la}{\leftarrow}

\def\A{{\cal A}}
\def\B{{\cal B}}
\def\C{{\cal C}}
\def\D{{\cal D}}
\def\E{{\cal E}}
\def\F{{\cal F}}
\def\G{{\cal G}}
\def\H{{\cal H}}
\def\I{{\cal I}}
\def\J{{\cal J}}
\def\K{{\cal K}}
\def\L{{\cal L}}
\def\M{{\cal M}}
\def\N{{\cal N}}
\def\O{{\cal O}}
\def\P{{\cal P}}
\def\Q{{\cal Q}}
\def\R{{\cal R}}
\def\S{{\cal S}}
\def\T{{\cal T}}
\def\U{{\cal U}}
\def\V{{\cal V}}
\def\W{{\cal W}}
\def\X{{\cal X}}
\def\Y{{\cal Y}}
\def\Z{{\cal Z}}
\def\Re{{\mathbb R}}
\def\Cx{{\mathbb C}}
\def\Ze{{\mathbb Z}}
\def\Na{{\mathbb N}}
\def\ud{\mathrm{d}}
\def\eps{\varepsilon}
\def\dist{\textrm{dist}}

\definecolor{VangelisColor}{rgb}{0,0,0.8} 
\newcommand{\kalo}[1]{{\color{VangelisColor} \textbf{[Vangelis: #1]}}}
\definecolor{MariosColor}{rgb}{0.8,0,0} 
\newcommand{\marios}[1]{{\color{MariosColor} \textbf{Marios: #1}}}

%% file: content/sections/abstract.tex
\begin{abstract}
We present a deep learning method
that propagates point-wise feature representations across shapes within a collection for the purpose of 3D shape segmentation. 
We propose a cross-shape attention mechanism to enable interactions between a shape's point-wise features and those of other shapes.
The mechanism assesses both the degree of interaction between points and also mediates feature propagation across shapes,
improving the accuracy and consistency of the resulting point-wise feature representations for shape segmentation. 
Our method also proposes a shape retrieval measure to select suitable shapes for cross-shape attention operations for each test shape. 
 Our experiments demonstrate that our approach yields state-of-the-art results in the popular PartNet dataset.

\ccsdesc[500]{Computing methodologies~Shape representations}
\ccsdesc[500]{Computer systems organization~Neural networks}

\printccsdesc
\end{abstract}

%% file: content/sections/introduction.tex
\vspace*{-10mm}
\section{Introduction}
\label{sec:intro}
\vspace*{-1mm}
Learning effective point-based representations
is fundamental to shape understanding and processing. In recent years, there has been 
significant research in developing deep neural
architectures to learn point-wise representations of shapes through convolution and attention 
layers, useful for performing 
high-level tasks, such as shape segmentation. The common denominator of these networks is that they output a representation for each shape point by weighting and aggregating representations and relations with
other points within the same shape.

In this work, we propose a \emph{cross-shape attention} mechanism that enables interaction and propagation of point-wise feature representations 
across shapes of an input collection. In our architecture, the representation of a point in a shape is learned by combining
representations originating from points in the same shape as well as other shapes. The rationale for such an approach is that if a point on one
shape is related to a point on another shape e.g., they lie on geometrically or semantically similar patches or parts, then cross-shape 
attention can promote consistency in their resulting representations and part label assignments. We leverage neural attention to 
determine and weigh pairs of points on different shapes. We integrate these weights in our cross-shape attention scheme to learn more consistent point representations for the purpose of semantic shape segmentation. 

Developing such a cross-shape attention mechanism is challenging. Performing cross-attention across all pairs of shapes becomes prohibitively  expensive   for large input collections of shapes. Our method learns a measure that allows us to select a small set of other shapes useful for such cross-attention operations with a given input shape. For example, given an input office chair, it is more useful to allow interactions of its points with points of another structurally similar office chair rather than a stool.
During training, we maintain a sparse graph (Figure \ref{fig:csn_teaser}),
whose nodes represent training shapes and edges specify which pairs of shapes should interact for training our cross-shape attention mechanism. At test time, the shape collection graph is augmented with additional nodes representing test shapes. New edges are added connecting them to training shapes for propagating representations from relevant training shapes.

We tested our cross-shape attention mechanism on two different backbones to extract the initial point-wise features per shape for the task of part segmentation: a sparse tensor network based on MinkowskiNet \cite{Choy:2019} and the octree-based network MID-FC \cite{Wang:2021}. For both backbones, we observed that our mechanism significantly improves the point-wise features for segmentation. Compared to the MinkowskiNet baseline, we found an improvement of $3.1\%$ in  mean part IoU in the PartNet benchmark \cite{mo2019partnet}. Compared to MID-FC, we found an improvement of $+1.3\%$  in  mean part IoU, achieving a new state-of-the-art result in PartNet (MID-FC: $60.8\%$ $\rightarrow$ Ours: $62.1\%$).

In summary, our main technical contribution is an attention-based mechanism that enables point-wise feature interaction and propagation within and across shapes for more consistent segmentation. Our experiments show state-of-the-art performance on the recent PartNet dataset.

%% file: content/sections/related_work.tex
\section{Related work}
\label{sec:related_work}

We briefly overview related work on 3D deep learning for point clouds. We also discuss cross-attention networks developed in other domains.  

\vspace*{-4mm}
\subsection{3D deep learning for processing point clouds} Several different types of neural networks have been proposed for processing point sets over the recent years. \rev{After the pioneering work of PointNet \cite{Qi17pointnet,Qi17}, several works further investigated more sophisticated point aggregation mechanisms to better model the spatial distribution of points 
\cite{Jiaxin:sonet,srivastava2019geometric,Le:deeper:2019,Liu:2020,deng2021vn,pointsetvoting,Luo_2022_CVPR,ZHAO2022108626,NEURIPS2022_9318763d}.} Alternatively, point clouds can  be projected onto local views 
\cite{su15mvcnn,qi2016volmv,kalogerakis2017shapepfcn,Huang:2017:LMVCNN} and processed as regular grids through image-based convolutional networks. Another line of work converts point representations into volumetric grids 
\cite{wu2015shapenets,maturana_iros_2015,dai2017scannet,rethage2018eccv,Shi_2019_CVPR,Liu2019:pointvoxelcnn} and processes them through 3D convolutions.
Instead of uniform grids, hierarchical space partitioning structures (e.g., kd-trees, octrees, lattices) can be used to define  regular convolutions 
\cite{Riegler2017OctNet,klokov2017escape,wang2017ocnn,Wang:2018:AOP,su18splatnet,Wang:2021}. Another type of networks incorporate point-wise convolution operators to directly process point clouds 
\cite{Li:pointcnn,Hua17,Xie18,liu2019densepoint,accv2018Groh,Atzmon18,Hermosilla18,Wang:parametricconv:18,xu2018spidercnn,Wu18:pointconv,Komarichev2019,Thomas19:kpconv}.
Alternatively, shapes can be treated as graphs by connecting each point to other points within neighborhoods in a feature space. Then graph convolution and pooling operations can be performed either in the spatial domain 
\cite{wang2018dynamic,shen2018mining,Davis18,wang2018local,zhang_linked_2019,Liu2019,Landrieu2019point,li2021deepgcns_pami,jiang2019hierarchical,xu2019gridgcn,Wang2019gcn,han2019point2node,Le:deeper:2019,SGPMarios},
or spectral domain \cite{Yi2017:syncspec,Boscaini2015spectral,Boscaini2016,Monti2017}.
Attention mechanisms have also been investigated to modulate the importance  of  graph edges and point-wise convolutions  \cite{Xie18,xu2019gridgcn,Wang2019gcn,shi2019pairwiseattention}. Graph neural network approaches have  been shown to model non-local interactions between points within the same shape \cite{wang2018dynamic,li2021deepgcns_pami,xu2019gridgcn,han2019point2node}. 
\rev{Finally, several recent works \cite{zhao2021point,guo2021pct, engel2021point, Mazur_2021_ICCV,yu2021pointbert,Xiang_2021_ICCV,pang2022masked,lai2022stratified,yang2023swin3d,Schult23ICRA}}
introduced a variety of transformer-inspired models for point cloud processing tasks. 
None of the above approaches have investigated the possibility of extending
attention across shapes. A notable exception are the methods by Wang et al. \cite{Wang_2019_ICCV} and Cao et al.  \cite{Cao_2021_ICCV} that propose cross-attention mechanisms across given pairs of point cloud instances representing different transformations of the same underlying shape for the specific task of rigid registration. 
Our method instead introduces cross-attention across shapes within a large collection without assuming any pre-specified shape pairs. Our method aims to discover useful pairs for cross-shape attention and learns representations by propagating them within the shape collection. Our method shows that the resulting features yield more consistent 3D shape segmentation than several other existing point-based networks. 

\vspace*{-4mm}
\subsection{Cross-attention in other domains} Our method is inspired by recent cross-attention models proposed for video classification, image classification, keypoint recognition, and image-text matching. Wang et al. \cite{wang2018non} introduced non-local networks that allow  any image query position to perceive features of all the other positions within the same image or across frames in a video. To avoid huge attention maps, Huang et al. proposes a ``criss-cross'' attention module \cite{huang2018ccnet} to maintain sparse connections for each position in image feature maps.
Cao et al. \cite{cao2019GCNet} 
 simplifies non-local blocks
with  query-independent attention maps \cite{cao2019GCNet}.
Lee et al. \cite{lee2018stacked} propose cross-attention between text and images to discover latent alignments between image regions and words in a sentence.
Hou et al. \cite{hou2019cross} models the semantic relevance between class and query feature maps in images through cross-attention to localize more relevant image regions for classification and generate more discriminative features.  Sarlin et al. \cite{sarlin2019superglue} learns keypoint matching between two indoor images from different viewpoints by leveraging self-attention and cross-attention to boost  the receptive field of local descriptors and allow cross-image communication. Chen et al. \cite{Chen_2021_ICCV} propose cross-attention between multiscape representations for image classification. Finally, Doersch et al. \cite{doersch2020crosstransformers} introduced a CrossTransformer model for few-shot learning on images. Give an unlabeled query image, their model computes local cross-attention similarities with a number of labeled images and then infers class membership.

Our method instead introduces attention mechanisms across 3D shapes. In contrast to cross-attention approaches in 
the above domains, we do not assume any pre-existing paired data. The usefulness of shape pairs is determined based
on a learned shape compatibility measure.

\begin{figure*}[!t]
    \centering
    \includegraphics[width=\textwidth]{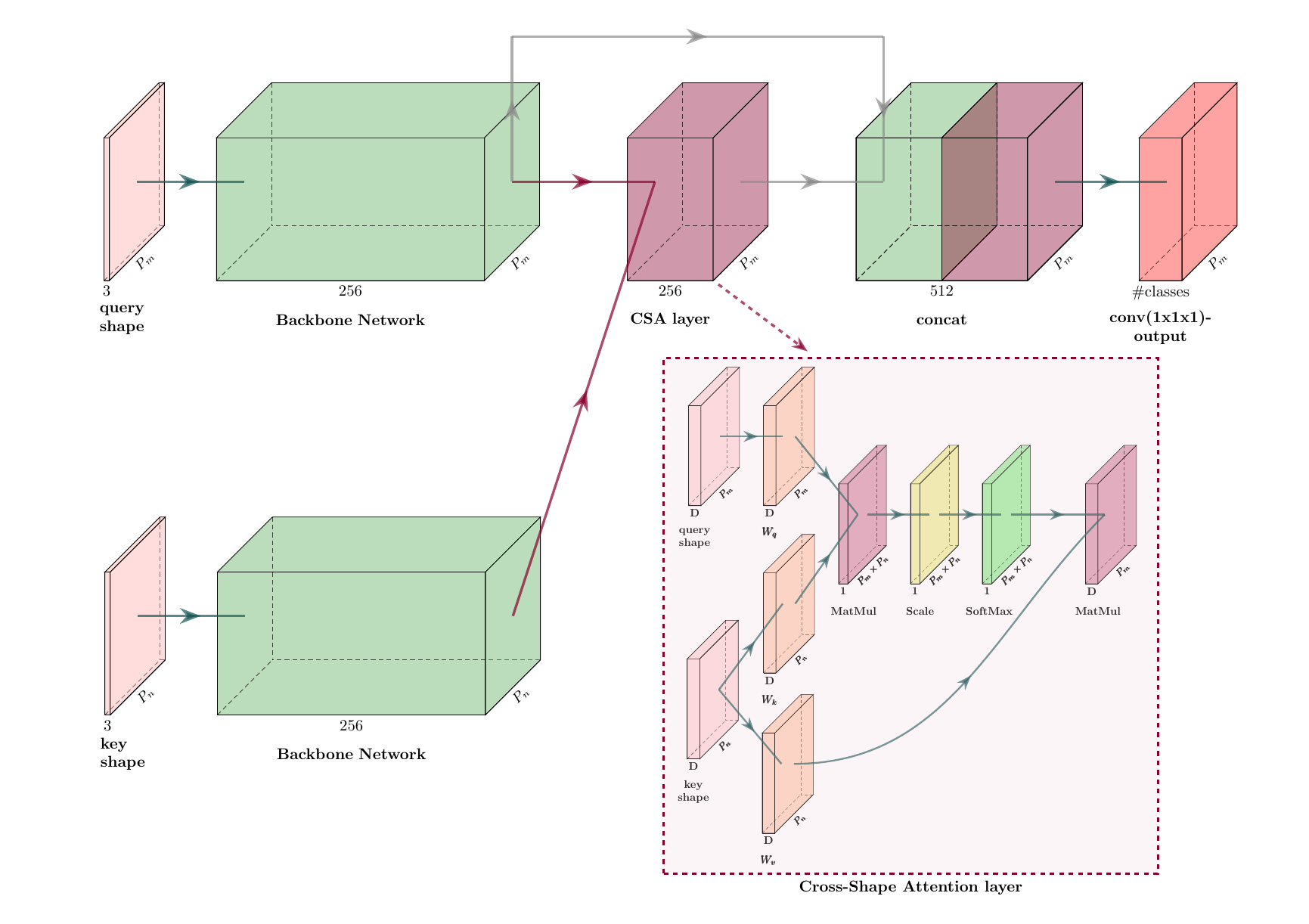}
    \vspace{-10mm}
    \caption{Our cross-shape network architecture: given an input test shape (``query shape'') represented as an input point set, we first extract initial point-wise features through a backbone (our MinkowskiNet variant, called ``MinkNetHRNet'', or alternatively the MID-FC network \cite{Wang:2021}). Then our proposed cross-attention layer, called CSA layer, propagates
    features extracted from another shape of the input shape collection (``key shape'') to the query shape such that their semantic segmentation becomes more synchronized. The output point-wise features of the CSA layer are concatenated with the original features of the query shape, then they are passed to a classification layer for semantic segmentation. Note that the illustrated CSA layer in the inlet figure uses only one head ($H=1$).}
    \label{fig:csn_architecture}
    \vspace{-5mm}
\end{figure*}

%% file: content/sections/method.tex
\vspace*{-2mm}
\section{Method}
\label{sec:method}

Given an input collection of 3D shapes represented as point clouds, the goal of our method is to extract and propagate point-based feature representations from one shape to
another, and use the resulting representations for 3D semantic segmentation. To perform the feature propagation, we propose a Cross-Shape Attention (CSA) mechanism. The mechanism first assesses the degree of interaction between pairs of points on different shapes. Then it
allows point-wise features on one shape to influence the point-wise features of the other shape based on their assessed degree of interaction. In addition, we  provide a mechanism that automatically selects shapes (``key shapes'') to pair with an input test shape (``query shape'') to execute these cross-shape attention operations. In the following sections, we first discuss the CSA layer at test time (Section \ref{subsec:cross_shape_attention}). Then we discuss our retrieval mechanism to find key shapes given a test shape (Section \ref{subsec:retrieval}), our training (Section \ref{subsec:training}),  test stage (Section \ref{subsec:test}),
and finally our network architecture details (Section \ref{subsec:csn_architecture}).

\vspace*{-3mm}
\subsection{Cross-shape attention for a shape pair}
\label{subsec:cross_shape_attention}

The input to our CSA layer is a pair of shapes represented as point clouds:
$ \mS_m = \{ \bp_i \}_{i=1}^{P_m} $
and 
$ \mS_n = \{ \bp_j \}_{j=1}^{P_n} $
where $\bp_i, \bp_j \in \mR^3$ represent 3D point positions and $P_m, P_n$ 
are the number of points for each shape respectively. Our first step is to extract point-wise features for each shape.

In our implementation, we experimented with two backbones for point-wise feature extraction: a sparse tensor network based on a modified version of MinkowskiNet 
\cite{Choy:2019}, and an octree-based network, called MID-FC \cite{Wang:2021} (architecture details for the two backbones are provided in Section
\ref{subsec:csn_architecture} and supplementary material). The output from the backbone is a per-point $D$-dimensional representation stacked into a matrix for each of the two shapes respectively: $\bX_m \in \mathcal{R}^{P_m \times D}$ and $\bX_n \in \mathcal{R}^{P_n\times D}$. The CSA layer produces new $D$-dimensional point-wise representations for both shapes:
\begin{equation}
  \pmb{X}'_m = f\big(\pmb{X}_m, \pmb{X}_n; \pmb{\theta}\big), \ \ \pmb{X}'_n = f\big(\pmb{X}_n, \pmb{X}_m; \pmb{\theta}\big)
\end{equation}
\noindent where $f$ is the cross-shape attention function with learned parameters $\theta$ described in the next paragraphs.

\vspace{-1.4mm}    
\paragraph*{Key and query intermediate representations.}
Inspired by  Transformers  \cite{Vaswani:2017}, we first transform the input point representations of the first shape in the pair
to intermediate representations, called ``query'' representations. The input point representations of the second shape are  transformed to 
intermediate ``key'' representations. The keys will be compared to queries to determine the degree of influence of one point on another. 
Specifically,  these transformations are expressed as follows:
\begin{equation}
    \pmb{q}_{m,i}^{(h)} = \pmb{W}_q^{(h)}
    \pmb{x}_{m,i}, \ \ \pmb{k}_{n,j}^{(h)} = \pmb{W}_k^{(h)}
    \pmb{x}_{n,j}
    \label{eq:key_query_repr}
\end{equation}
\noindent where $\pmb{x}_{m,i}$ and $\pmb{x}_{n,j}$ are point representations for the query shape $\mS_m$ and key shape $\mS_n$, $\pmb{W}_q^{(h)}$
and $\pmb{W}_k^{(h)}$ are $D'\times D$  learned transformation matrices shared across all points of the query and key shape respectively, and $h$ is an index denoting each different transformation (``head'').  The dimensionality of the key and query representations $D'$ is set to $\lfloor D/H \rfloor$, where $H$ is the number of heads. These intermediate representations are stacked into the matrices $\pmb{Q}_m^{(h)} \in 
\mathcal{R}^{P_m \times D'}$ and $\pmb{K}_n^{(h)} \in \mathcal{R}^{P_n \times D'}$. Furthermore, the point representations of the key shape $\mS_n$ are
transformed to value representations as:
\begin{equation}
    \pmb{v}_{n,j}^{(h)} = \pmb{W}_v^{(h)} \pmb{x}_{n,j}
    \label{eq:val_repr}
\end{equation}
\noindent where $\pmb{W}_v^{(h)}$ is a learned $D' \times D$ transformation shared across the points of the key shape. These are also stacked to a matrix $\pmb{V}_n^{(h)} \in \mathcal{R}^{P_n \times D'}$.

\vspace{-2.25mm}
\paragraph*{Pairwise point attention.}
The similarity of key and query representations is determined through scaled dot product \cite{Vaswani:2017}. This provides a measure of 
how much one shape point influences the point on the other shape. The similarity of key and query representations is determined for each head  as:
\begin{equation}
    \pmb{A}_{m,n}^{(h)} = softmax\Bigg(
    \frac{\pmb{Q}_m^{(h)} \cdot
    \big(\pmb{K}_n^{(h)}\big)^\top}{\sqrt{D'}}\Bigg)
    \label{eq:attention}
\end{equation}
\noindent where $\pmb{A}_{m,n}^{(h)} \in \mathcal{R}^{P_m \times P_n}$ is a  cross-attention matrix 
between the two shapes for each head.

\vspace{-1.75mm}
\paragraph*{Feature representation updates.}
The cross-attention matrix is used to update the point representations for the query shape $\mS_m$:
\begin{equation}
    \pmb{z}^{(h)}_{m,i} =  \sum_{j=1}^{P_n}\pmb{A}_{m,n}^{(h)}[i,j]\pmb{W}_v^{(h)}\pmb{x}_{n,j}
    \label{eq:cross_conv}
\end{equation}
The point-wise features are concatenated across all heads, then a linear transformation layer projects them back to  
$D$-dimensional space and they are added back to the original point-wise features of the query shape: 
\begin{equation}
\pmb{x}'_{m,i} = \pmb{x}_{m,i} + \bW_d \cdot [ \pmb{z}^{(1)}_{m,i}, \pmb{z}^{(2)}_{m,i}, ..., \pmb{z}^{(H)}_{m,i} ] 
\label{eq:final_conv}
\end{equation}
\noindent where $H$ is the number of heads and $\bW_d$ is another linear transformation. The features are stacked into a matrix $\pmb{X}'_m \in \mathcal{R}^{P_m \times D}$, followed by layer normalization \cite{Ba:2016LayerN}.

\vspace{-1.75mm}
\paragraph*{Self-shape attention.}
The pairwise attention of Equation \ref{eq:attention} and update operation of Equation \ref{eq:cross_conv} can also be 
applied to a pair that consists of the shape and itself. In this case, the CSA layer is equivalent to Self-Shape Attention (SSA), 
enabling long-range interactions between shape points within the same shape.

\vspace{-1.75mm}
\paragraph*{Cross-shape attention for multiple shapes.} 
We can further generalize the cross-shape operation in order to handle multiple shapes and  also combine it with
self-shape attention. Given a selected set of key shapes, our CSA layer outputs point representations for the query shape $\mS_m$ as follows:
\begin{equation}
    \pmb{X}'_m = \sum_{n \in \{\mathcal{C}(m), m\}} c(m, n) \pmb{A}_{m,n}\pmb{V}_n
    \label{eq:cross_conv_multiple_shapes}
\end{equation}
where $\mathcal{C}(m)$ is a set of key shapes deemed compatible for cross-shape attention with shape $\mS_m$ and $c(m,n)$
is a learned pairwise compatibility function between the query shape $\mS_m$ and each key shape $\mS_n$. The key idea of the above operation is to update point 
representations of the query shape $\mS_m$ as a weighted average of attention-modulated representations computed by using other key shapes as well as the 
shape itself. The compatibility function $c(m,n)$ assesses these weights that different shapes should have for cross-shape attention. It
also implicitly provides the weight of self-shape attention when $\mS_m = \mS_n$.

\vspace{-3mm}
\paragraph*{Compatibility function.}
To compute the compatibility function, we first extract a global, $D$-dimensional vector representation $\pmb{y}_m$ and $\pmb{y}_n$ for the query shape $\mS_m$ and each key shape $\mS_n$ 
respectively through mean-pooling on their self-shape attention representations:
\begin{equation}
    \pmb{y}^{(SSA)}_m = \mbox{avg}_i \pmb{X}'^{(SSA)}_{m,i} = \mbox{avg}_i \big(\pmb{A}_{m,m}\pmb{V}_m\big)
\end{equation}
\begin{equation}
    \pmb{y}^{(SSA)}_n = \mbox{avg}_i \pmb{X}'^{(SSA)}_{n,i} = \mbox{avg}_i \big(\pmb{A}_{n,n}\pmb{V}_n\big)
\end{equation}
In this manner, the self-attention representations of both shapes  provide cues for the compatibility between them expressed using their scaled dot product similarity
\cite{Vaswani:2017}:
\begin{align}
    \pmb{u}_m &= \pmb{U}_q\pmb{y}^{(SSA)}_m \nonumber \\ 
    \pmb{u}_n &= \pmb{U}_k\pmb{y}^{(SSA)}_n \nonumber \\
    s(m,n) &= \pmb{\hat{u}}_m \cdot \pmb{\hat{u}}_n^\top  
\label{eq:similarity} 
\end{align}
\noindent where $\pmb{U}_q$ and $\pmb{U}_k$ are learned $D \times D$ transformations for the query and key shape respectively, and 
\mbox{
$\pmb{\hat{u}}_m=
\pmb{{u}}_m / ||\pmb{{u}}_m|| 
,\pmb{\hat{u}}_n= 
\pmb{{u}}_n / ||\pmb{{u}}_n||$}. The final compatibility function $c(m,n)$ is computed as a normalized measure using a softmax transformation of compatibilities of the shape $m$ with all
other shapes in the set $\mathcal{C}(m)$, including the self-compatibility:
\begin{equation}
    c(m,n) = \frac{exp\big(s(m,n)\big)}{\sum_{n \in \{C(m), m\}}exp\big(s(m,n)\big)}
    \label{eq:compatibility}
\end{equation}
\subsection{Key shape retrieval}
\label{subsec:retrieval}
To perform cross-shape attention, we need to retrieve one or more key shapes for each query
shape. One possibility is to use the measure of Eq. \ref{eq:similarity} to evaluate the compatibility of the query shape with each candidate key shape from  an input collection. However, we found that this compatibility is 
more appropriate for the particular task of weighting the contribution of each selected key shape for cross-shape attention, rather than retrieving key shapes themselves (see supplementary for additional discussion). We instead found that it is  better to retrieve key shapes whose point-wise representations are on average more similar to the ones of the query shape. To achieve this, we perform the following steps: 

(i) We compute the similarity between points of the query shape and the points of candidate key shapes in terms of cosine similarity of their SSA representations:
\begin{equation}
\pmb{S}_{m,n} = \pmb{{X}}'^{(SSA)}_{m} \cdot ( \pmb{{X}}'^{(SSA)}_{n} )^\top
\end{equation}
\noindent where $\pmb{S}_{m,n} \in \mathcal{R}^{P_m \times P_n}$. 

(ii) Then for each query point, we find its best matching candidate key shape point yielding the highest cosine similarity:
\begin{equation}
r_i(m,n) = \max\limits_j \pmb{S}_{m,n}[i, j]
\vspace*{-2mm}
\end{equation}
(iii) Finally, we compute the average of these highest similarities across all query points:
\begin{equation}
r(m,n) = \mathop {avg}\limits_i r_i(m,n)
\label{eq:retrieval_measure}
\vspace*{-2mm}
\end{equation}
The retrieval measure $r(m,n)$ is used to compare the query shape $\mS_n$ with candidate key shapes from a collection.

\vspace*{-3mm}
\subsection{Training}
\label{subsec:training}
The input to our training procedure is a collection of point clouds with part annotations along with a smaller annotated collection
used for hold-out validation. We first train our backbone including a layer that implements self-shape attention alone according to Eq. \ref{eq:val_repr}-\ref{eq:final_conv} (i.e., $\mS_m = \mS_n$ in this case). The resulting output features are passed to a softmax layer for semantic segmentation. The network is trained according to softmax loss.
Based on the resulting SSA features, we construct a graph  (Figure \ref{fig:csn_teaser}), where 
 each training shape is connected with $K$  shapes, deemed as ``key'' shapes,
 according to our retrieval measure of Eq. \ref{eq:retrieval_measure}. One such graph is constructed for
the training split, and another for the  validation split. We then fine-tune our backbone and train
 a layer that implements our full cross-shape attention involving all $K$ key shapes per training shape using the same loss.  During training, we measure the performance of the network on the validation split, in terms of part IoU \cite{Mo:2019}, and if it reaches a plateau state, we recalculate the $K$-neighborhood of each shape based on the updated features. We further fine-tune our backbone and CSA layer. This iteration of graph update and fine-tuning of our network is performed two times in our implementation.

\vspace*{-4mm}
\subsection{Inference}
\label{subsec:inference}
\label{subsec:test}
\label{subsec:testing}
During inference, we create a graph connecting each test shape with $K$ training shapes retrieved by the measure of Eq. \ref{eq:retrieval_measure}. We then perform a feed-forward pass through our backbone, CSA layer, and classification layer to assess the label probabilities for each test shape.

\begin{figure*}[!t]
    \centering
    \includegraphics[width=0.95\textwidth]{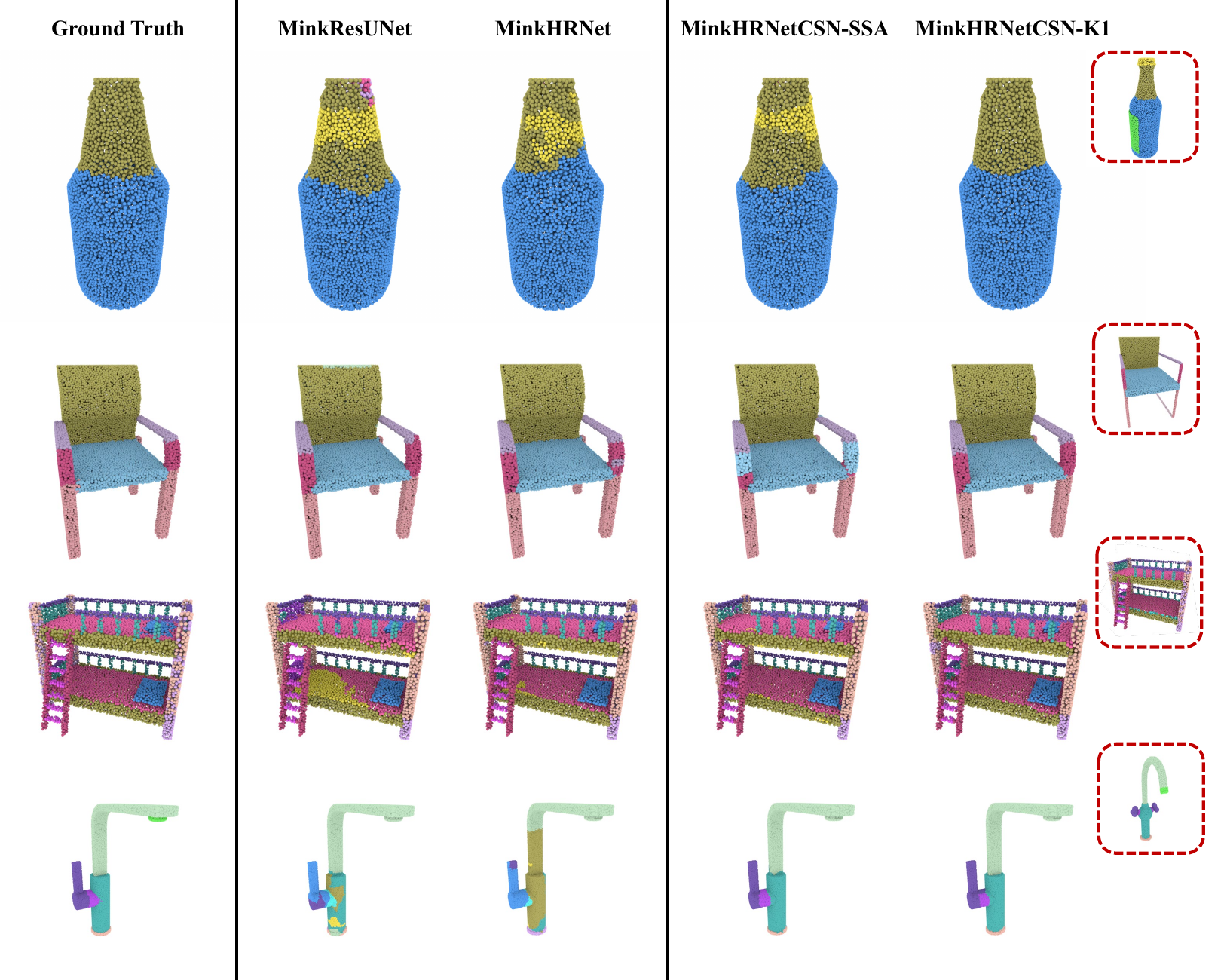}
    \vspace*{-3mm}
    \caption{Qualitative comparisons for a few characteristic test shapes of PartNet between the original MinkowskiNet for 3D shape segmentation (``MinkResUNet''), our backbone (``MinkHRNet''), and CrossShapeNet (CSN) in case of using self-shape attention alone (``MinkHRNetCSN-SSA'') and using cross-shape attention with $K=1$ key shape per query shape
    (``MinkHRNetCSN-K1''). The inlet images (red dotted box) show this key shape retrieved for each of the test shapes.
    }
    \label{fig:comparisons1}
\vspace{-6mm}     
\end{figure*}

\begin{figure*}[!t]
    \centering
\includegraphics[width=0.95\textwidth]{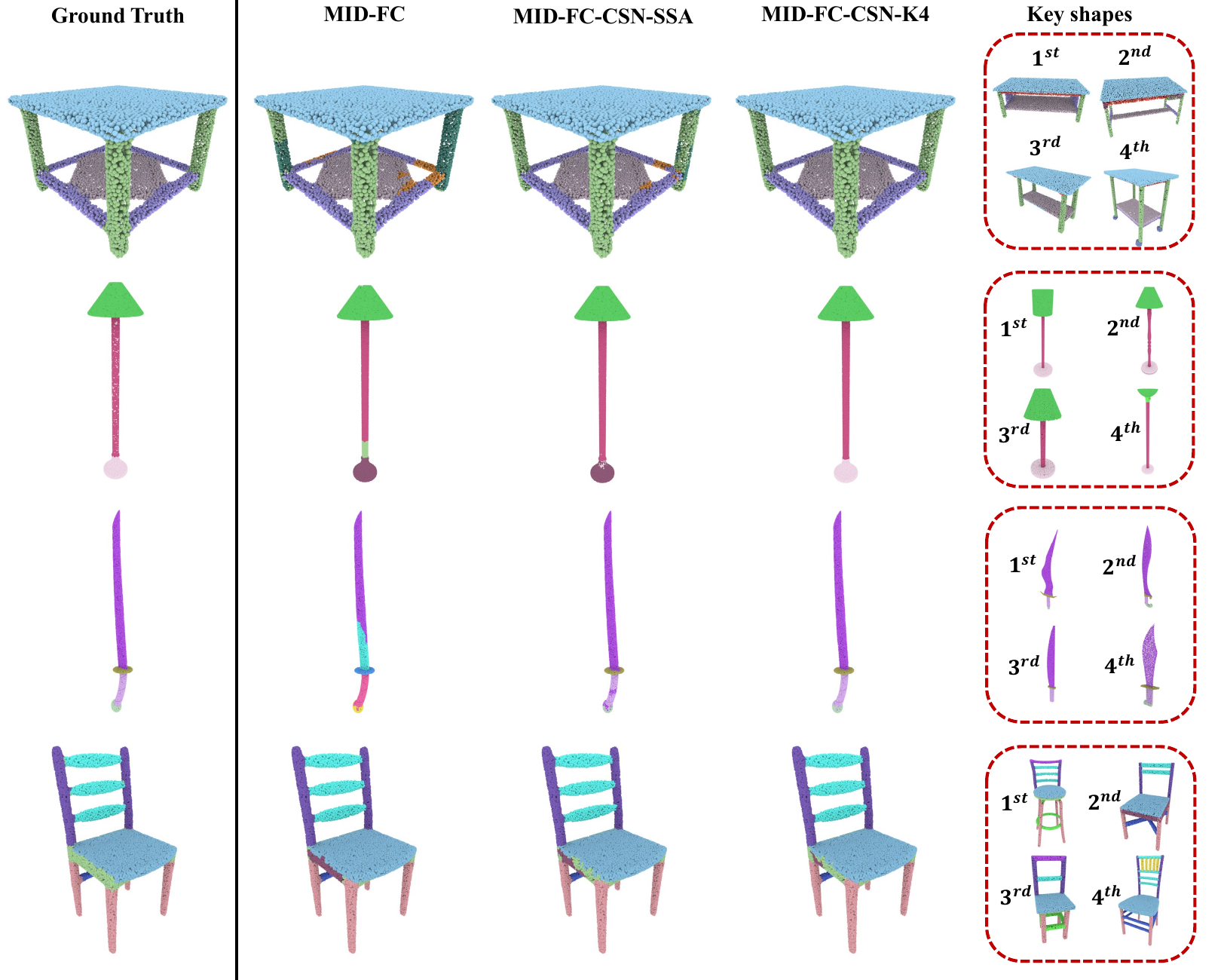}
    \vspace*{-4mm}
    \caption{Qualitative comparisons for a few characteristic test shapes of PartNet between the original MID-FC network for 3D shape segmentation (``MID-FC'') \cite{Wang:2021}, and CrossShapeNet (CSN) in case of using self-shape attention alone (``MID-FC-CSN-SSA'') and using cross-shape attention with $K=4$ key shape per query shape
    (``MID-FC-CSN-K4''). The last column shows the key shapes and their ordering, retrieved for each test shape.}
    \label{fig:comparisons2}
\vspace{-5mm}         
\end{figure*}

\begin{table}[!t]
    \begin{center}
    \begin{adjustbox}{max width=\textwidth}
        \begin{tabular}{*{20}{c}}
        \toprule
        Variant & \textbf{avg part IoU} \\
        \midrule
 MinkResUNet       & 46.8 \\ 
 MinkHRNet         & 48.0 \\
 MinkHRNetCSN-SSA  & 48.7 \\ 
 MinkHRNetCSN-K1   & \textbf{49.9} \\
 MinkHRNetCSN-K2   & 49.7 \\
 MinkHRNetCSN-K3   & 47.2 \\ 
\midrule
\midrule
 MID-FC            & 60.8  \\ 
 MID-FC-SSA        & 61.8  \\
 MID-FC-CSN-K1     & 61.9  \\ 
 MID-FC-CSN-K2     & 61.9  \\ 
 MID-FC-CSN-K3     & 62.0  \\ 
 MID-FC-CSN-K4     & \textbf{62.1}  \\ 
 MID-FC-CSN-K5     & 62.0  \\          \bottomrule
        \end{tabular}
    \end{adjustbox}
    \end{center}
    \vspace{-1mm}
    \caption{Ablation study for all our variants in PartNet.}
    \label{table:ablation}
\vspace*{-7mm}     
\end{table}

\vspace*{-3mm}
\subsection{Architecture}
\label{subsec:csn_architecture}
Here we describe the 
two backbones (MinkNetHRNet, MID-FC) we used to provide point-wise features to our CSA layer.

\vspace{-1.75mm}
\paragraph*{MinkNetHRNet.} The first backbone is a 
variant of the sparse tensor network based on MinkowskiNet \cite{Choy:2019}. 
We note that our variant performed better than the original MinkowskiNet  for 3D segmentation \cite{Choy:2019}, as discussed in our 
experiments. In a pre-processing step, we normalize the point clouds to a unit sphere and convert them to a sparse voxel grid (voxel size $=0.05$). After two convolutional layers, the network branches into three stages inspired by the HRNet \cite{HRNet:2019}, a network that processes 2D images in a multi-resolution manner. In our case, the first stage consists of three residual blocks processing the sparse voxel grid in its original resolution. The second stage downsamples the voxel grid by a factor of $2$ and processes it through two other residual blocks. The third stage further downsamples the voxel grid by a factor of $2$ and processes it through another residual block. The multi-resolution features from the three stages are combined into one feature map through upsampling following \cite{HRNet:2019}. 
The resulting feature map is further processed by a $1D$ convolutional block. The sparse voxel features are then mapped back to points as done in the original MinkowskiNet \cite{Choy:2019}. Details about the architecture of this backbone are provided in the supplementary. 

\vspace{-1.75mm}
\paragraph*{MID-FC.} The second variant utilizes an octree-based architecture based on the MID-FC 
network \cite{Wang:2021}. This network also incorporates a three-stage HRNet   \cite{HRNet:2019} to effectively maintain and merge multi-scale resolution feature maps. To implement this architecture, 
each point cloud is first converted into an octree representation with a resolution of $64^3$. To train 
this network, a self-supervised learning approach is employed using a multi-resolution instance 
discrimination pretext task with ShapeNetCore55 \cite{Wang:2021}. The training process involves 
two losses: a shape instance discrimination loss to classify augmented copies of each shape instance 
and a point instance discrimination loss to classify the same points on the augmented copies of a shape 
instance. This joint learning approach enables the network to acquire generic shape and point encodings 
that can be used for shape analysis tasks. Finally, the pre-trained network is combined with 
two fully-connected layers and our CSA layer. During training for our segmentation task, the HRNet is frozen, while we train only the two fully-connected layers and CSA layer for efficiency reasons. Details about the architecture of this backbone are provided in the supplementary. 

\subsection{Implementation details}
\label{subsec:implementation_details}
We train our Cross Shape Network  for each object category of 
PartNet \cite{Mo:2019} separately, using the standard cross entropy loss, for $200$ epochs. We set 
the batch size equal to 8 for all variants (SSA, K=1,2,3). For optimization we use the SGD optimizer
\cite{SGD:2016} with a learning rate of 0.5 and momentum $=0.9$. We scale learning rate by a factor 
of $0.5$, whenever the loss of the hold-out validation split saturates (patience $=10$ epochs, 
cooldown $=10$ epochs). For updating the shape graph for the training and validation split, we measure the performance of the validation split in terms of Part IoU. If it reaches a saturation point (patience $=10$ epochs, cooldown $=5$ epochs), we load the best model up to that moment, based on Part IoU performance, and update the graph for both splits. The graph is updated twice throughout our training procedure. For all layers we use batch normalization \cite{BN:2015} with momentum $=0.02$, except for the CSA module, where the layer normalization \cite{Ba:2016LayerN} is adopted. We also refer readers to our project page with source code for more details.\footnote{\emph{Our project page \href{https://marios2019.github.io/CSN/}{marios2019.github.io/CSN} includes our code and trained models}.}

%% file: content/sections/results.tex
\section{Results}
\label{sec:results}

 \begin{table*}[!ht]
    \begin{center}
    \begin{adjustbox}{max width=\textwidth}
        \begin{tabular}{*{20}{c}}
        \toprule
        \textbf{Category } & Bed & Bott & Chai & Cloc & Dish & Disp & Door & Ear & Fauc & Knif & Lamp & Micr & Frid & Stor & Tabl & Tras & Vase & \textbf{avg.} & \textbf{\#cat.} \\
        \midrule
 SpiderCNN \cite{xu2018spidercnn}        &  36.2  &  32.2  &  30.0  &  24.8  &  50.0  &  80.1  &  30.5  &  37.2  &  44.1  &  22.2  &  19.6  &  43.9  &  39.1  &  44.6  &  20.1  &  42.4  &  32.4  &  37.0  &  0 \\
 PointNet++ \cite{Qi17}       &  30.3  &  41.4  &  39.2  &  41.6  &  50.1  &  80.7  &  32.6  &  38.4  &  52.4  &  34.1  &  25.3  &  48.5  &  36.4  &  40.5  &  33.9  &  46.7  &  49.8  &  42.5  &  0 \\
 ResGCN-28 \cite{li2021deepgcns_pami}        &  35.9  &  49.3  &  41.1  &  33.8  &  56.2  &  81.0  &  31.1  &  45.8  &  52.8  &  44.5  &  23.1  &  51.8  &  34.9  &  47.2  &  33.6  &  50.8  &  54.2  &  45.1  &  0 \\
 PointCNN  \cite{Li:pointcnn}        &  41.9  &  41.8  &  43.9  &  36.3  &  58.7  &  82.5  &  37.8  &  48.9  &  60.5  &  34.1  &  20.1  &  58.2  &  42.9  &  49.4  &  21.3  &  53.1  &  58.9  &  46.5  &  0 \\
 CloserLook3D \cite{Liu:2020}     &  49.5  &  49.4  &  48.3  &  49.0  &  65.6  &  84.2  &  56.8  &  53.8  &  62.4  &  39.3  &  24.7  &  61.3  &  55.5  &  54.6  &  44.8  &  56.9  &  58.2  &  53.8  &  0 \\
 MinkResUNet  \cite{Choy:2019}     &  39.4  &  44.2  &  42.3  &  35.4  &  57.8  &  82.4  &  33.9  &  45.8  &  57.8  &  46.7  &  25.0  &  53.7  &  40.5  &  45.0  &  35.7  &  50.6  &  58.8  &  46.8  &  0 \\ 
 MinkHRNetCSN-K1 (ours)  &  42.1  &  54.0  &  42.5  &  42.9  &  58.2  &  83.2  &  43.5  &  51.5  &  59.4  &  47.8  &  27.9  &  57.4  &  43.7  &  46.2  &  36.8  &  51.5  &  60.0  &  49.9  &  0 \\
 MID-FC  \cite{Wang:2021}          &  51.6  &  56.5  &\textbf{55.7} &  55.3  &  75.6  &  91.3  &  56.6  &  53.8  &  64.6  &  55.4  &  31.2  &  78.7  &  63.1  &  62.8  &  45.7  &  65.8  &  69.3  &  60.8  &  1 \\
 MID-FC-CSN-K4 (ours)    &\textbf{52.2} &\textbf{58.6} &\textbf{55.7} &\textbf{57.7} &\textbf{76.4} &\textbf{91.4} &\textbf{58.9} &\textbf{54.5} &\textbf{65.2} &\textbf{62.2} &\textbf{33.1} &\textbf{79.2} &\textbf{64.0} &\textbf{62.9} &\textbf{46.0} &\textbf{67.2} &\textbf{69.9} &\textbf{62.1} & \textbf{16} \\
         \bottomrule
        \end{tabular}
    \end{adjustbox}
    \end{center}
    \vspace*{-1mm}
    \caption{Comparisons with other methods reporting performance in PartNet. The column ``avg.'' reports the mean Part IoU (averaged over all $17$ categories). The last column ``\#cat'' counts the number of categories that a method wins over others.}
    \label{table:comparisons}
\vspace{-7mm}    
\end{table*}

We evaluated our method for fine-grained shape segmentation qualitatively and quantitatively. In the next paragraphs, we discuss the used dataset, evaluation
metrics,  comparisons, and an analysis considering the computation time and size of our CSA layer.

\vspace{-1.75mm}
\paragraph*{Dataset.}
We use the PartNet dataset \cite{Mo:2019} for training and evaluating our method according to its provided training, validation, and
testing splits. Our evaluation focuses on the fine-grained level of semantic segmentation, which includes 17 out of the 24 object 
categories present in the PartNet dataset. We trained our network and competing variants separately 
for each object category.

\vspace{-1.75mm}
\paragraph*{Evaluation Metrics.} For evaluating the performance of our method and variants, we used the standard Part IoU metric, as also proposed in the PartNet benchmark
\cite{Mo:2019}.  The goal of our evaluation is to verify the hypothesis that our self-attention and cross-shape attention mechanisms yield better features for segmentation than the ones produced by any of the two original backbones on the task of semantic shape segmentation.

\vspace{-1.75mm}
\paragraph*{Ablation.} Table \ref{table:ablation} reports the mean part IoU performance averaged the PartNet's part categories for the original backbones (``MinkHRNet'') and (''MID-FC''). We first observe that our backbone variant ``MinkHRNet'' improves over the original ``MinkResUNet'' proposed in \cite{Choy:2019}, yielding an improvement of $1.2\%$ in mean Part IoU. Our variant based on self-shape attention alone (``MinkHRNetCSN-SSA'') further improves our backbone by  $0.7\%$ in  Part IoU. We further examined the performance of our cross-shape attention (CSA layer) tested in the variants ``MinkHRNetCSN-K1'', ``MinkHRNetCSN-K2'', and ``MinkHRNetCSN-K3'', where we use \mbox{$K=1,2,3$} key shapes per query shape. Our CrossShapeNet with $K=1$
(``MinkHRNetCSN-K1'') offers the best performance on average by improving Part IoU by another $1.2\%$ with respect to using self-shape attention alone. When using $K=2$ key shapes in cross-shape attention, the performance drops slightly ($-0.2\%$ in Part IoU on average) compared to using $K=1$, and drops even more when using $K=3$. Thus, for the MinkowskiNet variants, it appears that the optimal number of key shapes is $K=1$; we suspect that the performance drop for higher $K$ is due to the issue that the chance of retrieving shapes that are incompatible to the query shape is increased with larger numbers of retrieved key shapes.

We also observe improvements using the MID-FC backbone. Note that this backbone has higher performance than the MinkowskiNet variants due to its pretraining and fine-tuning strategies \cite{Wang:2021}. Our variant based on self-shape attention alone (``MID-FC-SSA'') further improves the original MID-FC backbone by  $1.0\%$ in mean Part IoU.  When using cross-shape attention, the optimal performance is achieved when using $K=4$ key shapes (``MID-FC-CSN-K4''), which improves 
Part IoU by another $0.3\%$ with respect to using self-shape attention alone. 
We note that the above improvements are quite stable -- by repeating all experiments $15$ times, the standard deviation of mean Part IoU is $\sigma = 0.03\%$. This means that the above differences are  significant -- even the improvement of $0.3\%$ of ``MID-FC-CSN-K4'' over ``MID-FC-SSA'' is of scale $10\sigma$. 

\vspace{-2.25mm} 
\paragraph*{Comparisons with other methods.}  Table \ref{table:comparisons} includes comparisons with other methods reporting their performance on PartNet per category
\cite{xu2018spidercnn,Qi17,li2021deepgcns_pami,Li:pointcnn,Liu:2020,Wang:2021}
along with our best performing variants (``MinkHRNetCSN-K1'' and ``MID-FC-CSN-K4''). 
\emph{Compared to the original MinkowskiNet (``MinkResUNet''), our 
``MinkHRNetCSN-K1'' variant achieves an improvement of $3.1\%$ in terms of mean Part IoU in PartNet}. \emph{Compared to ``MID-FC'', our best variant (``MID-FC-CSN-K4'') also offers a noticeable improvement of $1.3\%$ in mean Part IoU}. \emph{To the best of our knowledge, the result of our best variant represents the highest mean Part IoU performance achieved in the PartNet benchmark so far. As it can been in the last column of Table \ref{table:comparisons}, our method improves performance for $16$ out of $17$ categories.}

\vspace{-2.25mm}
\paragraph*{Qualitative Results.}
Figures \ref{fig:comparisons1} shows qualitative comparisons for MinkowskiNet-based variants -- specifically our best variant in this case using cross-shape attention with $K=1$, self-shape attention, our MinkNetHR backbone, and the original MinkowskiNet. 
Our backbone often improves the labeling relative to the original MinkowskiNet (e.g., see bed mattress, or monitor base). 
Our cross-shape attention tends to further improve upon fine-grained details in the segmentation e.g., see the top of the bottle, the armrests in the chair, and the bottom of the blade, pushing the segmentation to be more consistent with the retrieved key shape shown in the inlet images. Figure \ref{fig:comparisons2} shows comparisons for the MID-FC-based variants, including using cross-shape attention with $K=4$, self-shape attention, and the original MID-FC. We can drive similar conclusions -- our method improves the consistency of segmentation especially for fine-grained details e.g., the lower  bars of the table, the bottom of the lamp, the handle of the sword, and the sides of the seat. 

\vspace{-2.25mm}
\paragraph*{Number of parameters.} Our CSA layer adds a relatively small overhead in terms of number of parameters. The MID-FC backbone has $1.8M$ parameters, the MinkHRNet has $24.8M$ parameters, while the CSA layer adds $0.4M$ parameters.

\vspace{-2.25mm}
\paragraph*{Computation.}
\rev{The cross-shape operation (Eq. \ref{eq:cross_conv_multiple_shapes}) has linear time complexity wrt the number of used key shapes ($K$) and quadratic time complexity wrt the number of points per shape during both training and testing. To accelerate computation, subsets of points could be used as key points, as done in sparsified formulations of attention \cite{zaheer2020big,liu2022asset}.
The construction of the 
shape graph during training has quadratic time complexity wrt the number of the training shapes in the input collection. This is because 
the retrieval measure (Eq. \ref{eq:retrieval_measure}) must be evaluated for all pairs of training shapes in our current implementation. For example, our CSA layer for $K=1$ requires approximately $105$ 
hours on a NVidia $V100$ to train for the largest PartNet category ($3.5\times$ more compared to  training either backbone alone) due to the iterative graph construction and fine-tuning discussed in  
Section \ref{subsec:training}. 
A more efficient implementation could involve a more hierarchical approach e.g., perform clustering to select only a subset of training shapes as candidate key shapes for the retrieval measure. During inference, the time required per test shape exhibits linear complexity wrt the number of training shapes used for retrieval. In our experiments, testing time ranges from $0.7$ sec (``Dish'' class with the 
smallest number of shapes) to $7.8$ sec (``Table'' class with the largest number of shapes).}

%% file: content/sections/conclusion.tex
\vspace{-1.75mm}
\section{Conclusion}
\label{sec:conclusion}
We presented a method that enables interaction of point-wise features across different shapes in a collection. The interaction is mediated through a new cross-shape attention mechanism. Our experiments show improvements of this interaction in the case of fine-grained shape segmentation. 

\vspace{-2.75mm}
\paragraph*{Limitations.} \rev{The performance increase comes with a higher computational cost at training and test time. It would be interesting to explore if further performance gains can be achieved through
self-supervised pre-training \cite{PointContrast:2020,Wang:2021,prifit2022}
that could in turn guide our attention mechanism. Sparsifying the attention and accelerating the key shape retrieval mechanism would also be important to decrease the time complexity.
Another future research direction is to explore how to generalize the cross-shape attention mechanism from single shapes to entire scenes. 
}
\vspace{-2.75mm}
\paragraph*{Acknowledgments.} This project has received funding from Adobe Research and the EU
H2020 Research and Innovation Programme and the Republic of Cyprus through the Deputy Ministry of 
Research, Innovation and Digital Policy (GA 739578).

%% file: supplementary.tex
\vspace{5mm}
\begin{center}
\textbf{-- Supplementary Material --}
\end{center}

\input{content_supp/sections/architecture}

\input{content_supp/sections/key_shape_retrieval}

%% file: content_supp/sections/architecture.tex
\begin{table*}[tb!]
    \begin{center}
        \begin{tabular}{*{3}{c}}
        \toprule
        & \textbf{Cross-shape network architecture} & $\leftarrow$ CSN\big(\textit{query} $\mS_m$, \textit{key} $\mS_n$, \textit{\#classes} $K$\big) \\
        \midrule
        Index & Layer & Out \\
        \midrule
        1 & \textit{Input}: $\mS_m, \mS_n$ & $P_m \times 3, P_n \times 3$ \\
        2 & \textit{Mink-HRNet}\big($\mS_m$, $3$, $256$\big) & $P_m \times 256$ - query point repr. \\
        3 & \textit{Mink-HRNet}\big($\mS_n$, $3$, $256$\big) & $P_n \times 256$ - key point repr. \\
        4 & \textit{CSA}\big(Out(2), Out(2), 256, 4\big) & $P_m \times 256$ - query SSA repr. \\
        5 & \textit{CSA}\big(Out(3), Out(3), 256, 4\big) & $P_n \times 256$ - key SSA repr. \\
        6 & \textit{Linear-Q}\big(\textit{avg-pool}\big(Out(4)\big), $256, 256$\big) & $1 \times 256$ - query global repr. \\
        7 & \textit{Linear-K}\big(\textit{avg-pool}\big(Out(4)\big), $256, 256$\big) & $1 \times 256$ - query global repr. \\
        8 & \textit{Linear-K}\big(\textit{avg-pool}\big(Out(5)\big), $256, 256$\big) & $1 \times 256$ - key global repr. \\
        9 & \textit{ScaledDotProduct}\big(Out(6), Out(7)\big) & $1 \times 1$ - query-query similarity \\
        10 & \textit{ScaledDotProduct}\big(Out(6), Out(8)\big) & $1 \times 1$ - query-key similarity \\
        11 & \textit{Softmax}\big(Out(9), Out(10)\big) & $2 \times 1$ - compatibility \\
        12 & \textit{CSA}\big(Out(2), Out(3), 256, 4\big) & $P_m \times 256$ - query CSA repr. \\
        13 & Out(4) $*$ \textit{compatbility}[0] + Out(12) $*$ \textit{compatbility}[1] & $P_m \times 256$ - cross-shape attention \\
        14 & \textit{Softmax}\big(\textit{Conv}\big(\textit{ConCat}\big(Out(2), Out(13)\big), 512, $K$\big)\big) & $P_m \times K$ - per-point part label probabilities \\
        \bottomrule
        \end{tabular}
    \end{center}
    \vspace{-1mm}
    \caption{Cross-shape network architecture for $K=1$ key shapes per query shape. }
    \label{table:csn_architecture}
\end{table*}

\begin{table*}[tb!]
    \begin{center}
    \begin{adjustbox}{max width=\textwidth}
        \begin{tabular}{*{20}{c}}
        \toprule
        \textbf{Category } & Bed & Bott & Chai & Cloc & Dish & Disp & Door & Ear & Fauc & Knif & Lamp & Micr & Frid & Stor & Tabl & Tras & Vase & \textbf{avg.} & \textbf{\#cat.} \\
        \midrule
        \midrule
        \textbf{Part IoU} \\
        \cmidrule{1-1}
        MinkHRNetCSN-K1 (Eq. 14) & \textbf{42.1} & \textbf{54.0} & \textbf{42.5} & \textbf{42.9} & \textbf{58.2} & \textbf{83.2} & \textbf{43.5} & \textbf{51.5} & \textbf{59.4} & \textbf{47.8} & \textbf{27.9} & \textbf{57.4} & 43.7 & \textbf{46.2} &       \textbf{36.8} & \textbf{51.5} & \textbf{60.0} & \textbf{49.9} & \textbf{16} \\
        MinkHRNetCSN-K1 (Eq. 10) & 38.7 & 47.0 & 41.9 & 40.8 & 55.7 & 82.3 & 41.3 & 50.5 & 57.9 & 37.3 & 24.7 & 56.2 & \textbf{44.1}        & 45.9 & 32.3 & 51.4 & 58.8 & 47.4 & 1 \\
        \midrule
        \midrule
        \end{tabular}
    \end{adjustbox}
    \end{center}
    \vspace{-1mm}
    \caption{Comparison of shape retrieval measures based on point-wise (Eq. 14, main text) and global (Eq. 10, main text) representations of a query-key pair of shapes, in terms of Part IoU and Shape IoU.}
    \label{table:shape_retrieval_ablation}
\end{table*}

\section*{Appendix A: Backbone architecture details}

\paragraph*{MinkHRNetCSN architecture details.} In Table \ref{table:csn_architecture} we describe the overall Cross-Shape Network architecture for $K=1$ key shapes per query shape, based on the HRNet \cite{HRNet:2019}
backbone (``MinkHRNetCSN-K1''). For $K=2,3$ and SSA variants, we use the same architecture. First, for an input query-key pair of shapes $\mS_m \in \mathcal{R}^{P_m \times 3}$ and $\mS_n \in \mathcal{R}^{P_n \times 3}$, point-wise
features $\pmb{X}_m$ and $\pmb{X}_n$ are extracted using the ``Mink-HRNet'' backbone (Layers 2 and 3). For each set of point features their self-shape attention representations are calculated via the \textit{Cross-Shape Attention} layer (Layers 4 and 5). The query shape point self-shape attention representations are then aggregated into a global feature using mean-pooling and undergo two separate linear transformations (\textit{Linear-Q} and \textit{Linear-K} in Layers 6 and 7, respectively). Leveraging these, the self-shape similarity is computed, using the \textit{scaled dot product} (Layer 9). For the
key shape point self-shape attention representations, we use only the \textit{Linear-K} transformation on the key shape's global feature (Layer 8), and calculate the query-key similarity (Layer 10). The compatibility for the cross-shape attention is computed as the softmax transformation of these two
similarity measures (Layer 11). The cross-shape point representations of the query shape, propagating point features from the key shape, are extracted by our CSA module in Layer 12. The self-shape (Layer 4) and cross-shape (Layer 12) point representations are combined together, weighted by the pairwise compatibility, resulting in the cross-shape attention representations $\pmb{X}'_m$ (Layer 13). Finally, part label probabilities are extracted per point, through a $1\times1\times1$ convolution and a softmax transformation (Layer 14), based on the concatenation of the query shape's backbone representations $\pmb{X}_m$ and cross-shape attention representations $\pmb{X}'_m$.

The architecture of our backbone network, ``Mink-HRNet'', is described in Table \ref{table:hrnet_backbone}. Based on an input shape, our backbone first extracts point representations through two consecutive convolutions (Layers 2-5). Then, three multi-resolution branches are deployed. The first branch, called \textit{High-ResNetBlock} (Layers 6, 8 and 15), operates on the input shape's resolution, while the other two, \textit{Mid-ResNetBlock} (Layers 9 and 16) and \textit{Low-ResNetBlock} (Layer 17), downsample the shape's resolution by a factor of 2 and 4, respectively. In addition, feature representations are exchanged between these branches, through downsampling and upsampling modules (Layers 7, 10-14). The point representations of the two low-resolution branches are upsampled to the original resolution (Layers 18-20) and by concatenating them with point features of the high-resolution branch, point representations are extracted for the
input shape, through a full-connected layer (Layers 21 and 22).

The \textit{Cross-Shape Attention} (CSA), \textit{Downsampling} and \textit{Upsampling} layers,
along with \textit{Residual Basic Block} are described in more detail in Table \ref{table:csn_basic_layers}.

\begin{table*}[tb!]
    \begin{center}
        \begin{tabular}{*{3}{c}}
        \toprule
        & \textbf{MID-FC-Cross-shape network architecture} & $\leftarrow$ MID-FC-CSN\big(\textit{query} $\mS_m$, \textit{key} $\mS_n$, \textit{\#classes} $K$\big) \\
        \midrule
        Index & Layer & Out \\
        \midrule
        1 & \textit{Input}: $\mS_m, \mS_n$ & $P_m \times 3, P_n \times 3$ \\
        2 & \textit{MID-Net}\big($\mS_m$, $3$, $256$\big) & $P_m \times 256$ - query point repr. \\
        3 & \textit{MID-Net}\big($\mS_n$, $3$, $256$\big) & $P_n \times 256$ - key point repr. \\
        4 & \textit{CSA}\big(Out(2), Out(2), 256, 8\big) & $P_m \times 256$ - query SSA repr. \\
        5 & \textit{CSA}\big(Out(3), Out(3), 256, 8\big) & $P_n \times 256$ - key SSA repr. \\
        6 & \textit{Linear-Q}\big(\textit{avg-pool}\big(Out(4)\big), $256, 256$\big) & $1 \times 256$ - query global repr. \\
        7 & \textit{Linear-K}\big(\textit{avg-pool}\big(Out(4)\big), $256, 256$\big) & $1 \times 256$ - query global repr. \\
        8 & \textit{Linear-K}\big(\textit{avg-pool}\big(Out(5)\big), $256, 256$\big) & $1 \times 256$ - key global repr. \\
        9 & \textit{ScaledDotProduct}\big(Out(6), Out(7)\big) & $1 \times 1$ - query-query similarity \\
        10 & \textit{ScaledDotProduct}\big(Out(6), Out(8)\big) & $1 \times 1$ - query-key similarity \\
        11 & \textit{Softmax}\big(Out(9), Out(10)\big) & $2 \times 1$ - compatibility \\
        12 & \textit{CSA}\big(Out(2), Out(3), 256, 8\big) & $P_m \times 256$ - query CSA repr. \\
        13 & Out(4) $*$ \textit{compatbility}[0] + Out(12) $*$ \textit{compatbility}[1] & $P_m \times 256$ - cross-shape attention \\
        14 & \textit{Softmax}\big(\textit{FC}\big(Out(13), 256, $K$\big)\big) & $P_m \times K$ - per-point part label probabilities \\
        \bottomrule
        \end{tabular}
    \end{center}
    \vspace{-1mm}
    \caption{MID-FC-CSN architecture for $K=1$ key shapes per query shape. }
    \vspace{-2mm}\label{table:mid_fc_csn_architecture}
\end{table*}

\begin{table*}[tb!]
    \begin{center}
    \begin{adjustbox}{max width=1\textwidth}
        \begin{tabular}{*{3}{c}}
        \toprule
        & \textbf{Mink-HRNet backbone} & $\leftarrow$ Mink-HRNet\big(\textit{shape repr.} $\pmb{X}_m$, \textit{in\_feat} $D_{in}$, \textit{out\_feat} $D_{out}$\big) \\
        \midrule
        Index & Layer & Out \\
        \midrule
        1 & \textit{Input}: $\pmb{X}_m$ & $P_m \times D_{in}$ \\
        2 & \textit{Conv}\big($\pmb{X}_m$, $D_{in}$, $32$\big) & $P_m \times 32$ \\
        3 & \textit{ReLU}\big(\textit{BatchNorm}\big(Out(2)\big)\big) & $P_m \times 32$ \\
        4 & \textit{Conv}\big(Out(3), $32$, $64$\big) & $P_m \times 64$ \\
        5 & \textit{ReLU}\big(\textit{BatchNorm}\big(Out(4)\big)\big) & $P_m \times 64$ \\
        6 & \textit{High-ResNetBlock}\big($3\times$ \textit{BasicBlock}\big(Out(5), $64$\big)\big) & $P_m \times64$ \\
        7 & \textit{Downsampling}\big(Out(6), 64, 128\big) & $P_m/2 \times 128$ \\
        8 & \textit{High-ResNetBlock}\big($3\times$ \textit{BasicBlock}\big(Out(6), $64$\big)\big) & $P_m \times64$ \\
        9 & \textit{Mid-ResNetBlock}\big($3\times$ \textit{BasicBlock}\big(\textit{ReLU}\big(Out(7)\big), $128$\big)\big) & $P_m/2 \times128$ \\
        10 & \textit{Downsampling}\big(Out(8), 64, 128\big) & $P_m/2 \times 128$ \\
        11 & \textit{ReLU}\big(\textit{Downsampling}\big(Out(8), 64, 128\big)\big) & $P_m/2 \times 128$ \\
        12 & \textit{Downsampling}\big(Out(11), 128, 256\big) & $P_m/4 \times 256$ \\
        13 & \textit{Upsampling}\big(Out(9), 128, 64\big) & $P_m \times 64$ \\
        14 & \textit{Downsampling}\big(Out(9), 128, 256\big)\big) & $P_m/4 \times 256$ \\
        15 & \textit{High-ResNetBlock}\big($3\times$ \textit{BasicBlock}\big(\textit{ReLU}\big(Out(8)+Out(13)\big), $64$\big)\big) & $P_m \times64$ \\
        16 & \textit{Mid-ResNetBlock}\big($3\times$ \textit{BasicBlock}\big(\textit{ReLU}\big(Out(9) + Out(10)\big), $128$\big)\big) & $P_m/2 \times128$ \\
        17 & \textit{Low-ResNetBlock}\big($3\times$ \textit{BasicBlock}\big(\textit{ReLU}\big(Out(12) + Out(14)\big), $256$\big)\big) & $P_m/4 \times256$ \\
        18 & \textit{ReLU}\big(\textit{Upsampling}\big(Out(16), 128, 128\big)\big) & $P_m \times 128$ \\
        19 & \textit{ReLU}\big(\textit{Upsampling}\big(Out(17), 256, 256\big)\big) & $P_m/2 \times 256$ \\
        20 & \textit{ReLU}\big(\textit{Upsampling}\big(Out(19), 256, 256\big)\big) & $P_m \times 256$ \\
        21 & \textit{Conv}\big(\textit{ConCat}\big(Out(3), Out(15), Out(18), Out(20)\big), 480, $D_{out}$\big) & $P_m \times D_{out}$ \\
        22 & \textit{ReLU}\big(\textit{BatchNorm}\big(Out(21)\big)\big) & $P_m \times D_{out}$ \\
        \bottomrule
        \end{tabular}
    \end{adjustbox}
    \end{center}
    \vspace{-1mm}
    \caption{Mink-HRNet backbone architecture. High, mid and low-resolution ResNet blocks consist of 3 consecutive residual basic blocks, each. Point representations are exchanged between multi-resolution branches via downsampling and upsampling layers (see Table \ref{table:csn_basic_layers} for a more detailed description of their architecture). The convolution kernel of Layer 2 is of size $5\times5\times5$, in order to increase its
    receptive field, while for Layer 4 is of size $3\times3\times3$. For Layer 21 we used a kernel of $1\times1\times1$, since this acts as a fully-connected layer.}
    \label{table:hrnet_backbone}
\end{table*}

 \begin{table*}[tb!]
    \begin{center}
    \begin{adjustbox}{max width=1\textwidth}
        \begin{tabular}{*{3}{c}}
        \toprule
        & \textbf{Cross-Shape Attention Layer} & $\leftarrow$ CSA\big(\textit{query} $\pmb{X}_m$, \textit{key} $\pmb{X}_n$, \textit{\#feats} $D$, \textit{\#heads} $H$\big) \\
        \midrule
        Index & Layer & Out \\
        \midrule
        1 & \textit{Input}: $\pmb{X}_m, \pmb{X}_n$ & $P_m \times D, P_n \times D$ \\
        2 & $H\times$ \textit{Linear-Q}\big($\pmb{X}_m, D, \lfloor D/H \rfloor$\big) & $P_m \times H \times D'$ \\
        3 & $H\times$ \textit{Linear-K}\big($\pmb{X}_n, D, \lfloor D/H \rfloor$\big) & $P_n \times H \times D'$ \\
        4 & $H\times$ \textit{Linear-V}\big($\pmb{X}_n, D, \lfloor D/H \rfloor$\big) & $P_n \times H \times D'$ \\
        5 & \textit{Attention}\big(Out(2), Out(3)\big) & $H \times P_m \times P_n$ \\
        6 & \textit{MatMul}\big(Out(5), Out(4)\big) & $P_m \times H \times D'$ \\
        7 & \textit{Linear}\big(\textit{ConCat}\big(Out(6)\big), $D$, $D$\big) & $P_m \times D$ \\
        8 & \textit{LayerNorm}\big($\pmb{X}_m + $ Out(7)\big) & $P_m \times D$ \\
        \midrule
        \midrule
        & \textbf{Downsampling Layer} & $\leftarrow$ Downsampling\big(\textit{shape repr.} $\pmb{X}_m$, \textit{in\_feat} $D_{in}$, \textit{out\_feat} $D_{out}$\big) \\
        \midrule
        1 & \textit{Input}: $\pmb{X}_m$ & $P_m \times D_{in}$ \\
        2 & \textit{Conv}\big($\pmb{X}_m$, $D_{in}$, $D_{out}$, \textit{stride} $=2$\big) & $P_m/2 \times D_{out}$ \\
        3 & \textit{BatchNorm}\big(Out(2)\big) & $P_m/2 \times D_{out}$ \\
        \midrule
        \midrule
        & \textbf{Upsampling Layer} & $\leftarrow$ Upsampling\big(\textit{shape repr.} $\pmb{X}_m$, \textit{in\_feat} $D_{in}$, \textit{out\_feat} $D_{out}$\big) \\
        \midrule
        1 & \textit{Input}: $\pmb{X}_m$ & $P_m \times D_{in}$ \\
        2 & \textit{TrConv}\big($\pmb{X}_m$, $D_{in}$, $D_{out}$, \textit{stride} $=2$\big) & $2*P_m \times D_{out}$ \\
        3 & \textit{BatchNorm}\big(Out(2)\big) & $2*P_m \times D_{out}$ \\
        \midrule
        \midrule
        & \textbf{Residual Basic Block} & $\leftarrow$ BasicBlock\big(\textit{shape repr.} $\pmb{X}_m$, \textit{\#feats} $D$\big) \\
        \midrule
        1 & \textit{Input}: $\pmb{X}_m$ & $P_m \times D$ \\
        2 & \textit{Conv}\big($\pmb{X}_m$, $D$, $D$\big) & $P_m \times D$ \\
        3 & \textit{ReLU}\big(\textit{BatchNorm}\big(Out(2)\big)\big) & $P_m \times D$ \\
        4 & \textit{Conv}\big(Out(3), $D$, $D$\big) & $P_m \times D$ \\
        5 & \textit{ReLU}\big($\pmb{X}_m$ + \textit{BatchNorm}\big(Out(4)\big)\big) & $P_m \times D$ \\
        \bottomrule
        \end{tabular}
    \end{adjustbox}
    \end{center}
    \vspace{-1mm}
    \caption{Cross-Shape Network basic layers. All convolution kernels are of size $3\times3\times3$.}
    \label{table:csn_basic_layers}
    \vspace{-2mm}
\end{table*}

\paragraph*{MID-FC-CSN architecture details.} Similar to the MinkHRNetCSN, the MID-FC-CSN variant also follows 
a comparable architecture (see Table \ref{table:mid_fc_csn_architecture}). To extract point features 
$\pmb{X}_m$ and $\pmb{X}_n$ for the input query-key pair of shapes, the ``MID-Net'' backbone is 
utilized (Layers 2 and 3). This backbone also adopts a three-stage HRNet architecture, which is built on 
an octree-based CNN framework \cite{wang2017ocnn}. ResNet blocks with a bottleneck structure 
\cite{Kaiming:2016} are used in all multi-resolution branches, and feature sharing is achieved using 
downsample and upsample exchange blocks, implemented by max-pooling and tri-linear up-sampling, respectively. 
The CSA module (Layers 4 and 12) is employed to construct the self-shape and cross-shape attention features 
for the query shape. These are then weighted by the learned pairwise compatibility (Layers 4-11) and 
aggregated to generate the final cross-shape attention representations $\pmb{X}'_m$ (Layer 13). Part label 
probabilities are extracted per point using a fully-connected layer and a softmax transformation based on the 
cross-shape attention representations (Layer 14).

%% file: content_supp/sections/key_shape_retrieval.tex
\section*{Appendix B: Key shape retrieval measure comparison}

As an additional ablation, we evaluated the performance of our 
``MinkHRNetCSN-K1'' variant for two key shape retrieval measures (see Section 3.2 in the main text). 
The first relies on the point-wise  representations between a query and a key shape and retrieves key
shapes that are on average  more similar to their query counterparts (Eq. \ref{eq:retrieval_measure}). 
The second  measure, takes into account only the global representations of a query-key pair of shapes
(Eq. \ref{eq:similarity}). In Table \ref{table:shape_retrieval_ablation} we report the performance for
both measures, in terms of Part IoU and Shape IoU. Our default variant, ``MinkHRNetCSN-K1 (Eq. 14)'',
achieves  better performance according to Part IoU ($+2.5\%$), and it outperforms the other variant (``MinkHRNetCSN-K1, Eq. 10)'' in 16 out 17 object 
categories. This is a strong indication that the key shape retrieval
measure based in Eq. 14 is more effective in retrieving key shapes for
cross-shape attention.